\documentclass[runningheads]{llncs}
\usepackage[T1]{fontenc}
\usepackage{graphicx}
\usepackage{amsfonts}
\usepackage{amsmath}
\usepackage{multirow}
\usepackage{booktabs}
\usepackage{tablefootnote}
\usepackage{threeparttable}
\usepackage{hyperref}
\usepackage{url}
\usepackage[misc]{ifsym}
\usepackage{algorithm}
\usepackage{algorithmic}
\usepackage{enumitem}
\usepackage{tabularx}

\begin{document}

%

\title{FedJudge: Federated Legal Large \\ Language Model}
\author{Linan Yue, Qi Liu$^{(\textrm{\Letter})}$, Yichao Du, Weibo Gao, Ye Liu, Fangzhou Yao}
\authorrunning{L. Yue et al.}
\institute{State Key Laboratory of Cognitive Intelligence, \\University of Science and Technology of China
\email{\{lnyue,duyichao,weibogao,liuyer,fangzhouyao\}@mail.ustc.edu.cn}; \\
\email{qiliuql@ustc.edu.cn} }
%
%

%
\maketitle              
\begin{abstract}
  Large Language Models (LLMs) have gained prominence in the field of Legal Intelligence, offering potential applications in assisting legal professionals and laymen. However, the centralized training of these Legal LLMs raises data privacy concerns, as legal data is distributed among various institutions containing sensitive individual information. This paper addresses this challenge by exploring the integration of Legal LLMs with Federated Learning (FL) methodologies. By employing FL, Legal LLMs can be fine-tuned locally on devices or clients, and their parameters are aggregated and distributed on a central server, ensuring data privacy without directly sharing raw data. 
  However, computation and communication overheads hinder the full fine-tuning of LLMs under the FL setting.
 Moreover, the distribution shift of legal data reduces the effectiveness of FL methods. To this end, in this paper, we propose the \textit{first} Federated Legal Large Language Model (FedJudge) framework, which fine-tunes Legal LLMs efficiently  and  effectively.
 Specifically,
FedJudge utilizes parameter-efficient fine-tuning methods to update only a few additional parameters during the FL training. Besides, we explore the continual learning methods to preserve the global model's important parameters when training local clients to mitigate the problem of data shifts.
Extensive experimental results on three real-world datasets clearly validate the effectiveness of FedJudge.
Code is released at \url{https://github.com/yuelinan/FedJudge}.

\keywords{Federated Learning  \and Domain-specific Large Language Model \and Parameter-efficient  Fine-tuning \and Continual Learning}

\end{abstract}
\section{Introduction}
\label{sec:intro}
\begin{figure}[!htp]
  \centering 

  \includegraphics[width = 9cm]{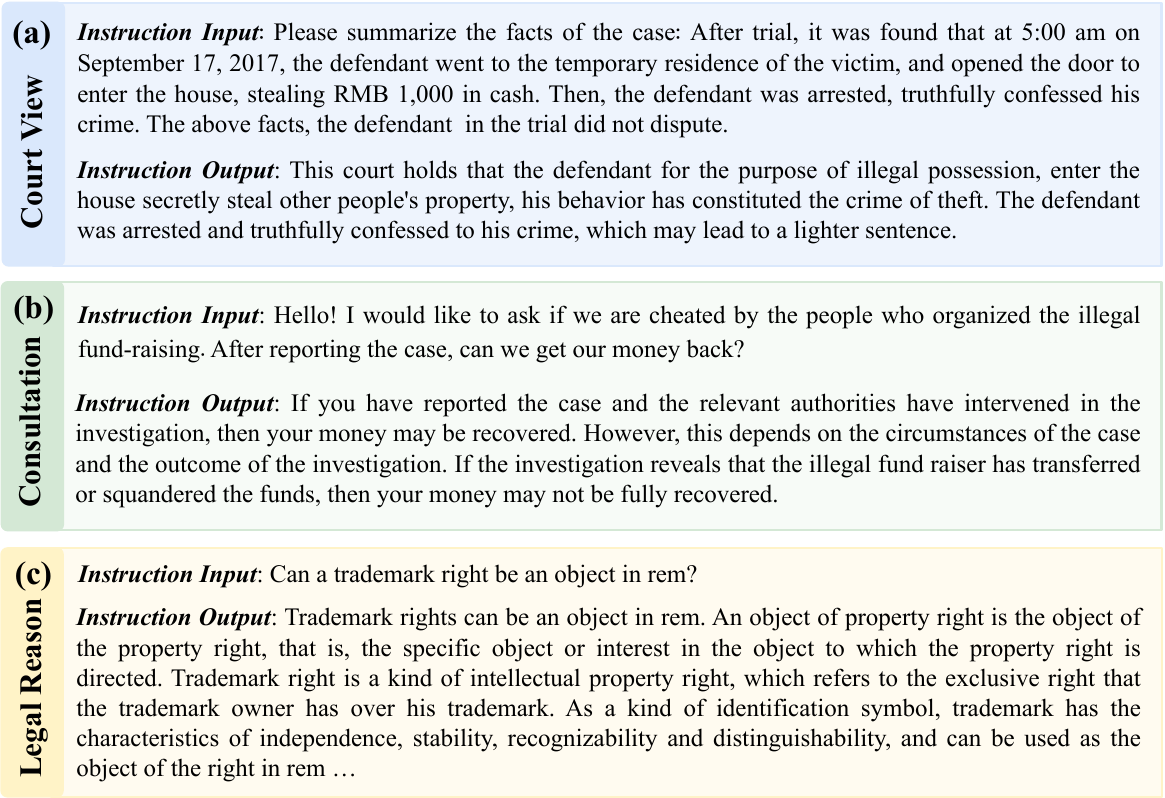}
 \caption{
  (a) An example in the court client. (b) An example in the legal consultation client. (c) An example in the legal reasoning client. From the figure, we can observe that the distribution of data from different clients varies significantly.
 }
 \label{example}

\end{figure}
Recently, the excellence of Large Language Models (LLMs) \cite{ChatGPT,touvron2023llama} promotes its application in the field of Legal Intelligence. 
By fine-tuning the base LLMs with instruction tuning on legal data,
numerous Legal LLMs have been proposed (e.g. Lawyer LLaMA \cite{lawyer-llama-report} and ChatLaw \cite{cui2023chatlaw}), which can assist legal professionals (e.g. lawyers and judges) to improve their work efficiency and provide legal consulting services to laymen.

Although Legal LLMs have achieved promising results, since they are trained on  centralized datasets, they still suffer from the data privacy problem. Specifically, considering a practical scenario, a huge amount of legal data is distributed among different institutions (e.g. courts, prosecutors' offices, and legal consulting firms) which all contain sensitive data of individuals.
Therefore, direct data communication can be impossible in reality, rendering the data-centric Legal LLMs training paradigms unavailable \cite{LuH0023,chen2023federated,zhang2023fedlegal}.

To this end, in this paper, we focus on how to fine-tune LLMs to the legal domain under the Federated Learning (FL) \cite{mcmahan2017communication,yang2019federated} setting.
By implementing the FL approach, Legal LLMs can be fine-tuned on local devices or on clients (e.g., courts and legal consulting firms clients). Then, Legal LLMs are updated by aggregating and distributing parameters at a central server, avoiding the sharing of raw data and effectively protecting the data privacy.

Although the prospects for integrating FL into Legal LLMs are promising, there is still a non-trivial question that requires answering:
\textit{How to efficiently and effectively fine-tune Legal LLMs with federated learning methods?}

Firstly, due to the huge number of parameters of LLM, fully fine-tuning LLMs is extremely resource-intensive and will bring vast computational overheads, which restricts the fine-tuning of several clients with limited computational capabilities.
Moreover, employing traditional FL algorithms (e.g., FedAvg \cite{mcmahan2017communication}) to aggregate and distribute the LLMs parameters increases the communication overhead in the FL system, rendering the standard FL workflow unavailable.
Both of them decrease the efficiency of  fine-tuning LLMs.

Besides, the distributional shift in legal data also affects the fine-tuning process of Large Language Models (LLMs). Figure \ref{example} provides an illustration of the data in the court, legal consulting and legal reasoning clients.
In court client, the text is typically presented in a legal professional linguistic style, characterized by formal and specialized language. The data from legal consulting clients often exhibits a more colloquial and informal style of description. Meanwhile, in the legal reasoning client, the data is more inclined to rigorous written descriptions.
This heterogeneity in the data poses challenges during training, leading to suboptimal aggregation performance and subsequently diminishing the effectiveness of Federated Learning (FL) methods.

To address the above problems, in this paper, we propose the \textit{first} Federated Legal Large Language Model (FedJudge) framework which fine-tunes LLMs to the legal domain efficiently and effectively.
Specifically, we propose to employ parameter-efficient fine-tuning \cite{zhang-etal-2023-fedpetuning} methods to fine-tune LLMs under the FL settings. We first adopt the LoRA \cite{hu2022lora} approach to train each local client separately, updating only the trainable rank decomposition matrices while  freezing LLMs weights.
Then, only trained LoRA parameters are uploaded to the central server and aggregated. Finally, the aggregated global LoRA parameters are distributed to each client to achieve efficient fine-tuning of the federated Legal~LLMs.

Moreover, 
the global model can be considered to have relatively minor distributional differences when compared to the local model.
Therefore, to address the problem of data distribution shifts, when updating local models, we constrain the local LoRA parameters so that they do not forget the important  global LoRA parameters with continual learning methods \cite{Zhou_2021_CVPR}.
Since the weight divergence between the local and global models is reduced and the knowledge of the global model is preserved, the problem of data shifts is mitigated.

In summary, the major contributions of this paper are shown as follows:
\begin{itemize}
  \item FedJudge is the \textit{first} federated Legal LLMs framework which considers both the computation and communication overheads in LLMs fine-tuning and the performance degradation due to legal data heterogeneity.
  \item We propose a parameter-efficient fine-tuning method for LoRA under the federated learning setting. Meanwhile, a continual learning method is introduced to prevent important knowledge of the global model from being forgotten during local model training.
  \item Extensive experiments on three legal tasks, including court view generation, legal logical reasoning and legal consultation tasks, provide empirical evidence demonstrating the effectiveness of FedJudge.
\end{itemize}

\section{Related Work}
\subsection{Federated Learning}
Considering data security and privacy protection, Federated Learning (FL) algorithms \cite{mcmahan2017communication,yang2019federated} have emerged and gained significant traction in fields such as computer vision \cite{lim2020federated,LuH0023}, data mining \cite{chai2020secure}, and natural language processing \cite{zhang-etal-2023-fedpetuning,DuZWL0C23,du2024communication}.
These algorithms offer a notable departure from traditional centralized machine learning approaches that involve data aggregation on a central server. By enabling data to remain locally stored and processed, FL mitigates various privacy risks and associated costs. 
Among these FL algorithms, \cite{mcmahan2017communication} proposed FedAVG, which introduced a multi-round model-based interaction framework between the server and clients. In each round, clients first performed training on their respective local sensitive data and subsequently send the model updates to the server. The server then aggregated these local updates to construct an enhanced global model.
To address the challenge of distributional migration between client data, FedCL \cite{yao2020continual} proposed the incorporation of parameter importance into the regularized local loss function using Elastic Weight Consolidation (EWC) \cite{kirkpatrick2017overcoming}. Additionally, researchers have explored the combination of meta-learning \cite{jiang2019improving,fallah2020personalized} and transfer learning \cite{chen2020fedhealth} techniques with FL to minimize discrepancies between the local models and the global model, thereby mitigating issues related to data distribution shifts.

\subsection{Legal Artificial Intelligence and Large Language Model}
With the increasing availability of legal documents\footnote{\url{https://wenshu.court.gov.cn/}}, legal artificial intelligence  (AI) tasks have garnered significant attention, including legal judgment prediction \cite{yue2021neurjudge,zhang2023contrastive}, court view generation \cite{yue2021circumstances}, legal question and answer tasks \cite{duan2019cjrc,zhong2020jec} and  explainable legal AI \cite{yuedare,yue2023interventional}. However, these current legal AI tasks encounter two primary challenges. Firstly, due to the involvement of citizens' privacy, much of the legal data cannot be openly shared, limiting the broader application of legal AI. To tackle this issue, \cite{zhang2023fedlegal} introduced a real-world federated learning benchmark for Legal AI and conducted experiments on various legal tasks. Nonetheless, both federated and non-federated legal AI approaches tend to focus on specific tasks, lacking a unified paradigm, which presents another challenge for legal AI tasks. The emergence of Large Language Models (LLMs) offers new possibilities for addressing this challenge.

LLMs \cite{ChatGPT,touvron2023llama} have demonstrated remarkable performance across various complex tasks \cite{zhang2019interactive,zhang2023eatn} and have had a significant societal impact. To enhance the effectiveness of legal AI tasks, researchers are increasingly combining legal tasks with large models in the form of dialogues, aiming to unify multiple legal tasks.
Among them, Lawyer LLaMA \cite{lawyer-llama-report} underwent continual pretraining on a vast legal corpus to systematically learn legal knowledge. Subsequently, Lawyer LLaMA fine-tuned the model using legal instruction data, enabling it to apply legal knowledge to specific scenarios. Another approach, ChatLaw \cite{cui2023chatlaw}, explored a larger base model to enhance the logical reasoning capabilities of the legal model. Although these approaches have yielded promising results, concerns regarding privacy pertaining to legal data have hindered their further application. To effectively address this issue during LLMs training, one potential solution involves integrating federated learning into LLMs training \cite{chen2023federated}. To this end, in this paper, we propose the \textit{first} federated Legal LLMs framework.
\begin{figure}[!htp]
  \centering 

  \includegraphics[width = 8.8cm]{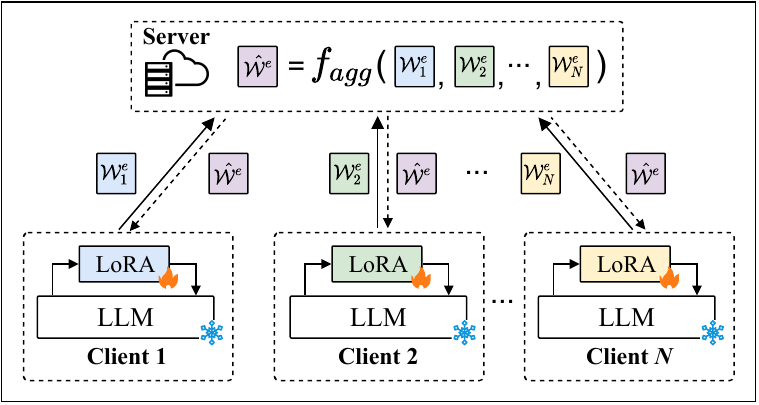}
  
 \caption{
  An overview of FedJudge, where \protect\includegraphics[width=7pt]{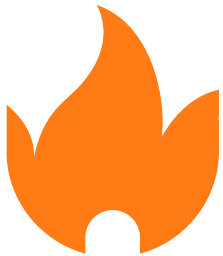} represents trainable weights and \protect\includegraphics[width=8pt]{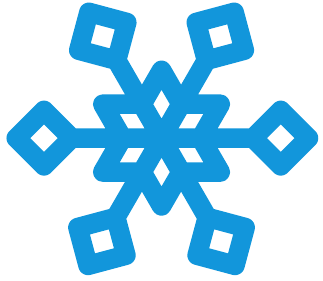} denotes frozen LLMs.
 }
 \label{model}
 
\end{figure}
\section{FedJudge: Federated Legal Large Language Model}
\label{sec:format}

\subsection{Problem Definition}

In the federated setting, we have $N$ clients $\left\{C_{1}, C_{2}, \cdots, C_{N}\right\}$, and their corresponding local private data $\left\{\mathcal{D}_{1}, \mathcal{D}_{2}, \cdots, \mathcal{D}_{N}\right\}$, where $P(\mathcal{D}_{i}) \neq P(\mathcal{D}_{j})$ (i.e., different clients have different data distributions).
In each client $C_{i}$, our goal is to employ the instruction tuning method to parameter-efficient fine-tune the base generative LLMs to the legal domain. The objective loss of autoregressive training is:
\begin{equation}
  \mathcal{L}_{i}=\max _{\mathcal{W}_{i}^{e}} \sum_{(x, y) \in \mathcal{D}_{i}} \sum_{m=1}^{|y|} \log \left({P}_{\mathcal{W}_{i}^{e}+\mathcal{W}_{i}^{p}}\left(y_{m} \mid x, y_{<m}\right)\right),
  \label{lora}
\end{equation}
where $x$ and $y$ represent the \textit{Instruction Input} and \textit{Instruction Output}, $y_{m}$ denotes the $m$-th token of $y$, $y_{<m}$ is the tokens before $y_{m}$, and $\mathcal{W}_{i}^{p}$ represents the frozen LLMs parameters and $\mathcal{W}_{i}^{e}$ is the trainable parameters ($\mathcal{W}_{i}^{e} \ll \mathcal{W}_{i}^{p}$).

Then, we only upload each $\mathcal{W}_{i}^{e}$ to the central server and use the aggregation function $f_{agg}(\cdot)$ to aggregate all clients parameters to yield the global parameters  $\hat{\mathcal{W}^{e}}$. Finally, we distribute $\hat{\mathcal{W}^{e}}$ to each client to complete one communication round of updating and training of FedJudge.

\subsection{Architecture of FedJudge}
In this section, we present the details of FedJudge. Firstly, to reduce the computation and communication overheads, we introduce the parameter-efficient fine-tuning method to FedJudge. This method is the backbone of FedJudge and is referred to FedJudge-Base.
Finally, based on the FedJudge-base, we explore the continual learning methods to alleviate the data distribution shifts problem in FL training, and we  denote it  as FedJudge-CL.


\begin{algorithm}[!htbp]
  \caption{Training process of FedJudge-CL} 
  \begin{algorithmic}
  \STATE \textbf{Server Executes:}
  \STATE Initialize the communication round $\mathcal{T}$, epoch $\mathcal{K}$, the numbers of clients $N$, and ${\mathcal{W}^{e}}^{(1)}$ on the server.
  \FOR{each communication round $t$=1 \textbf{to} $\mathcal{T}$}
  \FOR{each client id $i$=1 \textbf{to} $N$ \textbf{in parallel}}
  \STATE ClientUpdate($i$,${\mathcal{W}_{i}^{e}}^{(t)}$).
  \ENDFOR
  \STATE Receiving all local updated parameters:
  \STATE ${\mathcal{W}^{e}}^{(t)} = \left\{{\mathcal{W}_{1}^{e}}^{(t)}, {\mathcal{W}_{2}^{e}}^{(t)}, \cdots, {\mathcal{W}_{N}^{e}}^{(t)}\right\}$.
  \STATE Performing aggregation by  Eq.(\ref{agg}).

  \ENDFOR
  \STATE \textbf{ClientUpdate($i$,${\mathcal{W}_{i}^{e}}^{(t)}$):}

  \FOR{epoch $k$=1 \textbf{to} $\mathcal{K}$}
  \IF{t=1}
  \STATE Updating local parameters by Eq.(\ref{lora}).
  \ELSE
  \STATE $\delta {\mathcal{W}^{e}}^{(t)} = {\mathcal{W}_{i}^{e}}^{(t)} - {\hat{\mathcal{W}^{e}}}^{(t-1)}$.
    \STATE Calculating the Jacobian matrix$J^{(t-1)}$.
    \STATE Updating local parameters by Eq.(\ref{fedcl}).
  \ENDIF
  \ENDFOR

  \STATE Return local parameters ${\mathcal{W}_{i}^{e}}^{(t)}$ to server.
  \end{algorithmic}
  \label{algorithm}
  \end{algorithm}

\subsection{Parameter-Efficient Fine-Tuning in FedJudge}
 As shown in Figure \ref{model}, we introduce how to fine-tune the LLMs to the legal domain under the FL setting with parameter-efficient fine-tuning methods.
Inspired by \cite{zhang-etal-2023-fedpetuning}, we first employ the LoRA \cite{hu2022lora} method to train each local client $C_{i}$, where LLMs parameters are frozen and trainable rank decomposition matrices are introduced into each layer of the Transformer architecture \cite{vaswani2017attention} in LLMs.
The corresponding learning objective in each local client is shown in Eq.(\ref{lora}), where we only update $\mathcal{W}_{i}^{e}$ during the local training process.

After the clients are updated, we upload all clients LoRA parameters $\mathcal{W}^{e} = \left\{\mathcal{W}_{1}^{e}, \mathcal{W}_{2}^{e}, \cdots, \mathcal{W}_{N}^{e}\right\}$ to the central server. Then, we adopt the aggregation function $f_{agg}(\cdot)$ to aggregate all clients parameters. In this paper, we adopt the weighted average function as $f_{agg}(\cdot)$:
\begin{equation}
  \hat{\mathcal{W}^{e}}=\sum_{i=1}^{N} \frac{|D_{i}|}{\sum_{j=1}^{N} |D_{j}|} \mathcal{W}_{i}^{e}.
  \label{agg}
\end{equation}

For clarity, we rewrite $\mathcal{W}_{i}^{e}$ in the $(t)$-th communication round as ${\mathcal{W}_{i}^{e}}^{(t)}$ ($t \geq 1$) and $\mathcal{L}_{i}$ in Eq.(\ref{lora}) as ${\mathcal{L}_{i}}^{(t)}$.
At the end of the $(t)$-th round, we distribute  aggregated parameters ${\hat{\mathcal{W}^{e}}}^{(t)}$ to each client and replace the local ${\mathcal{W}^{e}}^{(t)}$ as  global~${\hat{\mathcal{W}^{e}}}^{(t)}$.

\subsection{Continual Learning in FedJudge}
Although FedJudge-Base (the backbone of FedJudge) can train FedJudge efficiently, it still suffers from the data distribution shifts problem, degrading the effectiveness of FedJudge. Therefore, in this subsection, we extend FedJudge-Base to FedJudge-CL  which explores the continual learning methods to alleviate the data shifts problem.
The core principle is that \textit{the more important the weight is, the less it is updated in the local training}.
Specifically, when training on local client, the local parameters which are distributed from the server are updated depending on parameter importance evaluated on the server. The more important the weight is, the less it is updated. The continuous learning method enables obtaining favorable performance on the local task while maintaining the performance on the global server. For instance, if the weight obtained from the server is evaluated as relatively unimportant, it is more frequently updated with increasing attention to improve the performance on the local task. On the contrary, the important weight obtained from the server is less possible to be updated in order to maintain the performance on the server. 

More specifically, we continue local training on each client based on the distributed parameters $\hat{\mathcal{W}^{e}}$. Then, we constrain important parameters in $\hat{\mathcal{W}^{e}}$ to change as slightly as possible during training.
This constraint can ensure local model does not forget the previous learned global knowledge.

To achieve this, in this paper, we employ the continual learning constraint in~\cite{Zhou_2021_CVPR}, where we define ${\hat{\mathcal{W}^{e}}}^{(t-1)}$ as the global parameters obtained and distributed in the ($t-1$)-th round of communication and ${\mathcal{W}_{i}^{e}}^{(t)}$ as the local trainable parameters for the ($t$)-th (current) round of training.
Specifically, we adopt the following equation to evaluate the change of parameters:
\begin{equation}
  {\mathcal{L}_{i}^{cl}}^{(t)} = {J^{(t-1)}}^{T}\left|\delta {\mathcal{W}^{e}}^{(t)}\right|+\left| \delta {\mathcal{W}^{e}}^{(t)}\right|^{T} {J^{(t-1)}}^{T} J^{(t-1)}\left| \delta {\mathcal{W}^{e}}^{(t)}\right|,
  \label{cl}
\end{equation}
where $\delta {\mathcal{W}^{e}}^{(t)} = {\mathcal{W}_{i}^{e}}^{(t)} - {\hat{\mathcal{W}^{e}}}^{(t-1)}$ indicates the change in parameter. $J^{(t-1)}$ is the Jacobian matrix of ${\hat{\mathcal{W}^{e}}}^{(t-1)}$ that represents the importance of the parameters.
Eq.(\ref{cl}) has been proven to be effective by \cite{Zhou_2021_CVPR}, and the detailed derivation can be referred to~\cite{Zhou_2021_CVPR}.
Finally, the objective of FedJudge-CL for each client $C_{i}$~is:
\begin{equation}
  {\mathcal{L}_{i}^{*}}^{(t)} = {\mathcal{L}_{i}}^{(t)} + \mathbb{I}(t \neq 1) \times  \lambda {\mathcal{L}_{i}^{cl}}^{(t)},
  \label{fedcl}
\end{equation}
where $\lambda$ is an adjustable parameter.
Detailed descriptions of FedJudge-CL is shown in Algorithm \ref{algorithm}.

\section{Experiments}
\label{sec:pagestyle}
To evaluate the effectiveness of FedJudge, we conduct experiments to answer the following research questions:
\begin{itemize}[leftmargin=*]
\item \textbf{RQ1:} How effective is FedJudge-Base and FedJudge-CL in improving the performance of legal AI  tasks under the FL scenario?
\begin{table*}
  \center
  \caption{The statistics of datasets.}
  \renewcommand\thetable{2}
  \setlength{\tabcolsep}{1.8mm}{
    \scalebox{.9}{
  \renewcommand\arraystretch{1.4}
  \begin{tabular}{cccc}
  \toprule
      Dataset  & Court View   & Consultation &  Reasoning\\
  \midrule
  \#Number of Training Set   & 47,303    & 13,000 & 6,000 \\
  \#Number of Test Set  &  12,624  & 2,797 & 1,000\\
  \#Avg.Instruction Input Length  & 480.1   &  57.3 & 135.8 \\
  \#Avg.Instruction Output Length  &  287.5   &  209.1  & 220.7\\
  \bottomrule
  \label{dataset}
\end{tabular}
    }}

\end{table*}

\item \textbf{RQ2:} Can the continual learning method alleviate  data shifts problems?
\item \textbf{RQ3:} How does FedJudge perform in human evaluation?
\item \textbf{RQ4:} What is the FedJudge's answer to a specific question?
\end{itemize}

\subsection{Datasets}

Under the FL setting, we employ three datasets with different distribution to evaluate the performance of FedJudge, including the court view generation dataset, legal consultation dataset and legal reasoning datasets. These datasets are not shared among the individual clients to simulate FL scenarios:
These three datasets are owned by three different clients (i.e. the court, consultation and reasoning clients), and the data is not shared among the individual clients to simulate FL scenarios.
Below, we present the details of the three datasets:

\begin{itemize}[leftmargin=*]
  \item \textbf{Court View Generation Dataset (Client1):} Court views are written to interpret verdicts and sentencing of the case fact by human judges. Therefore, in this dataset, our goal is to generate court views automatically based on the given case fact. To achieve this, we first collect 59,927 cases from the C3VG datasets. Then, following the instruction tuning methods \cite{alpaca}, we process the collected data into the form of \{\textit{Instruction Input}: \textit{Instruction Output}\}. 
  Among them, the \textit{Instruction Input} is: ``Assuming you are a judge, please  summarize the facts of the case: the description of case facts'', and  the \textit{Instruction Output} is the court views.
  Finally, we divide these data into the training and test set. Detailed statistics of datasets are shown in Table \ref{dataset}.
  Figure \ref{example}(a) shows an example in our processed~dataset.
  \item \textbf{Legal Consultation Dataset (Client2):} We first collect legal consultation data from Lawyer LLaMA \cite{lawyer-llama-report} as the training set, which presents the  consultation data  with the form of \{\textit{Instruction Input}: \textit{Instruction Output}\} naturally. Among them, \textit{Instruction Input} is a legal question asked by laymen in real-world scenarios, and \textit{Instruction Output} is generated by ChatGPT \cite{ChatGPT}.
  Then, we sample 2,797  consultation data from  a public dataset\footnote{\url{https://www.heywhale.com/mw/dataset/5e953ca8e7ec38002d02fca7/content}} as the test set.
  Figure~\ref{example}(b) shows an example of the consultation dataset.

  \begin{table*}[htbp]
  
    \centering
    
    \caption{Performance results of the FedJudge and baselines across three legal tasks. Among them, the \underline{underlined} scores represent the state-of-the-art (SOTA) scores in centralized training and the \textbf{bolded} are the SOTA scores under the FL setting.}
    
    \renewcommand\arraystretch{1.1}
  \setlength{\tabcolsep}{1.8mm}{
      \scalebox{0.71}{
        \begin{tabular}{c|cccccc|cccccccccccccccc}
      \hline
      \toprule
  
      \multirow{2}{*}{Methods} & \multicolumn{6}{c|}{\centering Court View Generation (Client1)} &  \multicolumn{6}{c}{\centering Legal Consultation (Client2)}\\
         & R-1 & R-2 & R-L& B-4  &B-N & Bert-S& R-1 & R-2 & R-L& B-4 &B-N & Bert-S\\
             \midrule
        Baichuan-7B &57.43 & 37.76 & 38.65 & 45.25 & 52.01 & 72.84 &
        37.10 & 10.65 & 14.82 & 14.83 & 19.76 & 62.33 \\
        \midrule
        Center &67.55 & 52.33 & 53.21 & 55.32 & 60.51 & 77.71 &
        \underline{38.93} & 11.76 & \underline{16.24} & 15.61 & 23.02 & 62.27 
        \\
        Center-Client1  &\underline{74.05} & \underline{60.25} & \underline{61.69} & \underline{60.13} & \underline{64.71} & \underline{82.58} &
        35.15 & \underline{12.44} & 15.11 & 11.89 & 16.37 & 62.02 \\
        Center-Client2 & 28.34 & 15.73 & 15.21 & 3.79 & 4.47 & 51.32 &
        38.61 & 12.22 & 15.94 & \underline{16.25} & \underline{23.70} & \underline{62.97} \\
        Center-Client3 & 39.22 & 19.31 & 18.91 & 19.76 & 24.83 & 61.67 &
        38.40 & 11.33 & 15.95 & 15.21 & 22.54 & 61.66 \\
        \midrule
            FedJudge-Base  &71.44 & 56.86 & 58.53 & 56.58 & 61.31 & 81.42 &
            39.66 & 12.74 & 16.44 & 12.23 & 17.52 & 63.48 \\
              Base-Client1 
              & 74.45 & 59.42 & 60.92 & 61.13 & \textbf{66.24} & 83.81 &
              36.82 & 12.63 & 15.58 & 9.42 & 13.06 & 62.20 
              \\
              Base-Client2 & 63.19 & 47.58 & 49.09 & 50.48 & 55.64 & 75.84 &
              39.83 & 11.92 & 16.17 & 14.46 & 21.41 & 64.09 \\
              Base-Client3 & 64.21 & 50.01 & 51.62 & 50.38 & 54.92 & 76.12 &
              {40.28} & 12.72 & 16.62 & 13.93 & 20.13 & 63.96 \\
              \midrule
              FedJudge-CL &72.41 & 57.57 & 59.10 & 59.14 & 64.14 & 82.09 &
              
              40.00 & \textbf{12.82} & 16.60 & 12.73 & 18.25 & 64.00 \\
              CL-Client1 &\textbf{74.82} & \textbf{59.57} & \textbf{61.13} & \textbf{61.46} & {61.46} & \textbf{84.18} &
              36.90 & 12.80 & 15.62 & 9.25 & 12.78 & 62.72 \\
              CL-Client2 &65.80 & 49.57 & 50.92 & 54.12 & 59.73 & 77.83 &          
              39.85 & 11.92 & \textbf{16.71} & \textbf{15.15} & \textbf{21.98} & \textbf{64.33}\\
              CL-Client3 & 67.19 & 52.28 & 53.92 & 53.56 & 58.45 & 78.45 &           
            \textbf{40.67} & 12.80  & 16.23 & 14.74 & 21.84 & 64.10 \\
      \bottomrule
      \hline
      \end{tabular}%
      }
      }   
      \label{table1}%
      
  \end{table*}

  \item \textbf{Legal Reasoning Dataset (Client3):} We also collect legal reasoning data from Lawyer LLaMA \cite{lawyer-llama-report}, where \textit{Instruction Input} is a question required fine-grained reasoning, and \textit{Instruction Output} is generated by ChatGPT \cite{ChatGPT} with detailed evidence and reasoning procedures. We also divide these data into the training and test set.
  Figure ~\ref{example}(c) shows an example of the reasoning dataset.
\end{itemize}

\subsection{Experimental Setup}
\subsubsection{Comparison methods}
In this paper, since we focus on Chinese law, we adopt Baichuan-7B \cite{baichuan}
as the pre-trained LLM backbone, which achieves competitive results in Chinese intelligence tasks.
Below, we introduce the baselines we use, which are also based on Baichuan-7B with LoRA:
\begin{itemize}[leftmargin=*]
\item  \textbf{Baichuan-7B}: We employ Baichuan-7B directly for prediction without instruction tuning.

\item \textbf{Center}: A standard centralized-training method which is trained with all data.

\item \textbf{Center-Client$\mathcal{E}$} ($\mathcal{E}=\{1,2,3\}$) : 
Centralized methods which are only trained on their private data in corresponding respective clients.

\item With FL methods, in addition to the global federated model FedJudge, we also obtain personalized models for the respective clients, which we refer to as \textbf{Base-Client$\mathcal{E}$} in FedJudge-Base, and \textbf{CL-Client$\mathcal{E}$} in FedJudge-CL.
\end{itemize}

\begin{table*}[htbp]
  
  \centering
  
  \caption{
  Performance results of the FedJudge and baselines on legal reasoning task.}
  
  \renewcommand\arraystretch{1.}
\setlength{\tabcolsep}{3.8mm}{
    \scalebox{0.78}{
      \begin{tabular}{c|cccccc}
    \hline
    \toprule
    \multirow{2}{*}{Methods} & 
    \multicolumn{6}{c}{\centering Legal Reasoning (Client3)}\\
       & R-1 & R-2 & R-L& B-4 &B-N & Bert-S\\
           \midrule
      Baichuan-7B &
      33.10 & 10.13 & 14.82 & 22.99 & 34.10 & 62.45 \\
      \midrule
      Center &
      \underline{57.09} & \underline{32.84} & \underline{35.02} & \underline{45.04} & \underline{54.98} & \underline{75.52}\\
      Center-Client1  &
      35.16 & 11.15 & 15.82 & 22.04 & 31.98 & 64.02\\
      Center-Client2 & 
      52.09 & 28.42 & 30.89 & 28.92 & 35.71 & 73.10\\
      Center-Client3 & 
      55.99 & 31.88 & 33.93 & 43.23 & 52.81 & 74.95\\
      \midrule
          FedJudge-Base  &
            53.55 & 28.33 & 31.16 & 40.96 & 51.40 & 73.50\\
            Base-Client1 &
            46.81 & 21.36 & 24.98 & 32.75 & 43.02 & 69.86 \\
            Base-Client2 &
            {55.14} & 30.25 & 32.81 & \textbf{42.50} & \textbf{52.74} & 74.44\\
            Base-Client3 & 
            54.60 & 29.61 & 32.24 & 41.91 & 52.18 & 74.16\\
            \midrule
            FedJudge-CL &
            53.36 & 27.94 & 30.92 & 41.31 & 52.16 & 73.32\\
            CL-Client1 &
            46.99 & 21.18 & 24.83 & 33.72 & 44.55 & 69.83\\
            CL-Client2 &\textbf{55.37} &29.72  &32.18 &42.03 &52.31 &74.17 \\
            CL-Client3 &
            {54.77} &
            \textbf{30.58} & \textbf{32.94} & 42.12 & 52.20 & \textbf{74.58} \\
           
    \bottomrule
    \hline
    \end{tabular}%
    }
    }   
    \label{table1-1}%

\end{table*}
\subsubsection{Evaluation Metrics}

We evaluate the effectiveness of FedJudge by employing conventional evaluation metrics, including ROUGE F1\footnote{We report the results of ROUGE-1, ROUGE-2, and ROUGE-L.}, BLEU\footnote{We calculate BLEU-4 and BLEU-N as evaluation metrics, where BLEU-N is the average of BLEU-1, BLEU2, BLEU-3 and BLEU-4.} scores and BertScore~\cite{ZhangKWWA20}.
Besides, we add a human evaluation to evaluate the text generated by FedJudge from two subjective metrics, including usefulness and fluency.

\subsubsection{Implementation Details}
In this paper, all methods including FedJudge and its variants utilize
Baichuan-7B \cite{baichuan} as the pre-trained LLM backbone.
We set the number of clients $N$ to 3, including the court view, legal consultation and reasoning clients.
During training, we set $\lambda$ in FedJudge-CL to 1.
The rank of LoRA is set to 4 and the communication round is set to 5.
The learning rate of the Adam optimizer \cite{KingmaB14} is initialized to 2e-4.
Experiments are performed on 2 Tesla A100 40G GPUs, where the batch size in each device is 2 and the gradient accumulation step is 8.

\subsection{Experimental Results}

\subsubsection{Overall Performance (RQ1).}
From Table \ref{table1} and Table \ref{table1-1}, we observe that Baichuan-7B without instruction tuning performs well on the legal task in the zero-shot setting, which indicates that Baichuan-7B has already possessed legal summarization and reasoning abilities through training on a large amount of data \cite{baichuan}. However, its results are still worse than the fine-tuned model (e.g., FedJudge-Base and FedJudge-CL), which also shows the necessity of fine-tuning LLMs to the legal domain.

\begin{figure}[!htp]
  \centering 

  \includegraphics[width = 10.cm]{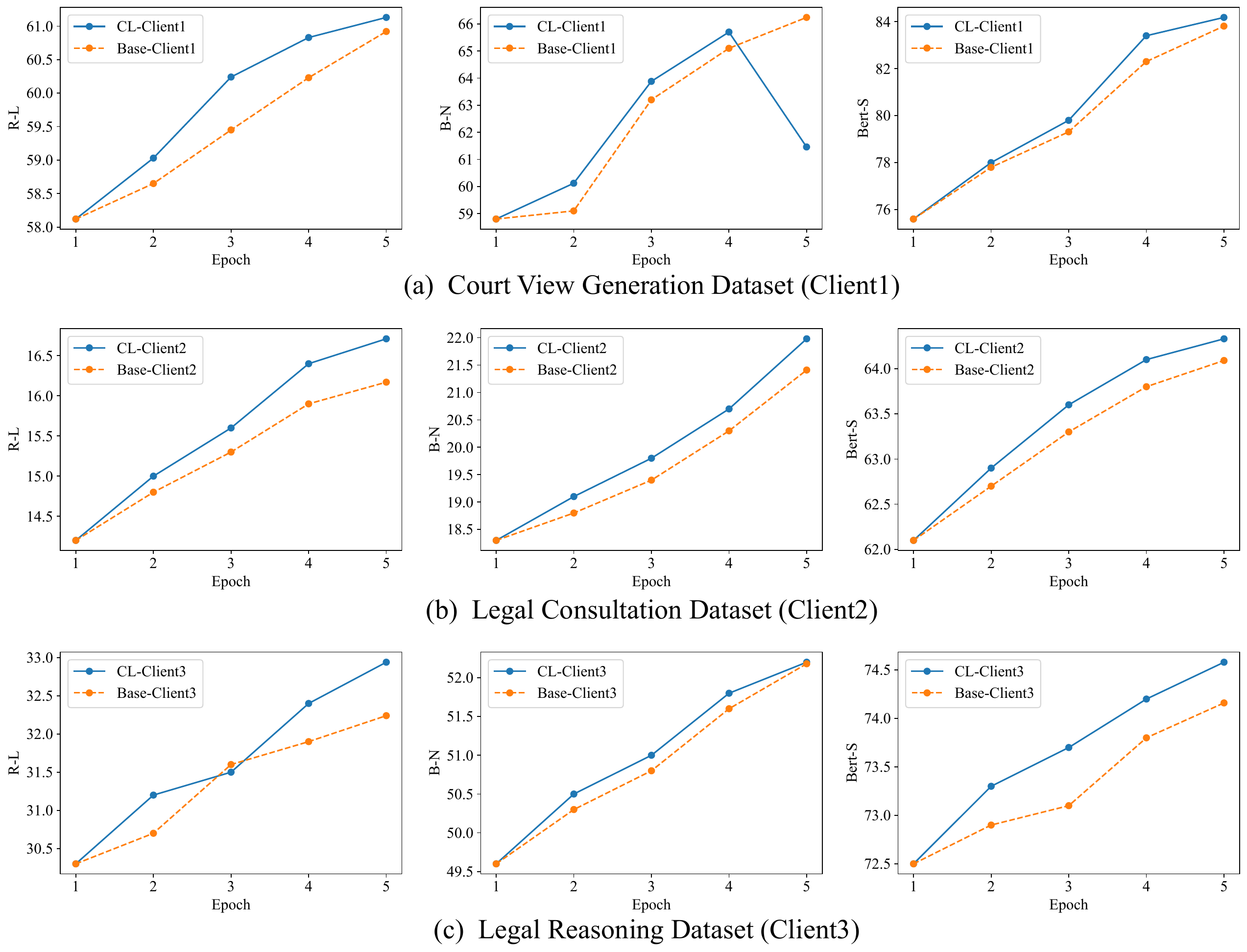}

 \caption{
  Training process of FedJudge on different clients.
 }
 
 \label{case3}

\end{figure}

Interestingly, although Center is a central model trained with all data, it still underperforms the federated model in several metrics, illustrating the unsuitability of simply mixing data with different distributions for centralized training.
Besides, Center-Client$\mathcal{E}$ obtained by training on the respective clients performs well on test data from the same distribution (e.g. Center-Client1 exceeds Center by 4.87~bertscore values on the client1 dataset.). However, it does not perform well on other data with different distributions (especially, Center-Client2 only achieves 4.47 BLEU-N values on the client1 data). This illustrates the imperative of using FL to obtain a global legal LLM in the context of data privacy.

We find FedJudge-Base achieves competitive results compared to Center, while our personalized model Base-Client$\mathcal{E}$ outperforms Center-Client$\mathcal{E}$, illustrating the effectiveness of fine-tuning LLMs with LoRA in the FL setting.
Finally, we observe that both FedJudge-CL and CL-Client$\mathcal{E}$ outperform FedJudge-Base and Base-Client$\mathcal{E}$ on most metrics,  suggesting the problem of data shifts can be mitigated by the continuous learning  constraints (Eq.(\ref{cl})). Meanwhile, as the local model is constrained not to forget the global knowledge, it facilitates the updating of the global model when performing Eq.(\ref{agg}) to obtain a more effective global model FedJudge-CL.

\subsubsection{Training Process of FedJudge (RQ2).}

To investigate the potential of continual learning in mitigating data shift problems, we present the changes in metrics (R-L, B-N, and Bert-S) for both CL-Client$\mathcal{E}$ and Base-Client$\mathcal{E}$ with respect to the training epochs, as shown in Figure \ref{case3}.
From the figure, we observe that in the initial epoch of training, when CL-Client$\mathcal{E}$ has not yet been trained using the continual learning method, both CL-Client$\mathcal{E}$ and Base-Client$\mathcal{E}$ exhibit the same metric values.
\begin{table*}
  \center
  \caption{Detailed scoring standards for human annotators.}
  \renewcommand\thetable{2}
  \setlength{\tabcolsep}{1.8mm}{
    \scalebox{.85}{
  \renewcommand\arraystretch{1.}
  \begin{tabular}{p{20pt}p{200pt}p{70pt}}
  \toprule
      Score  & Usefulness   & Fluency \\
      \midrule
      1 & No Use. The generated texts are useless for answering questions. & Nonsense.\\
      2 & Almost useless. Almost all generated texts are useless.& Very unfluent.\\
      3 & Half of them are useful. About half of the generated texts are useful for answering questions.  & Partial fluent.\\
      4 & Highly useful. Most generated texts are useful to answer the questions. & Highly fluent.\\
      5 & Exactly. Generated texts are useful for me to get the correct answer. & Very fluent.\\

  \bottomrule
  \label{human}
\end{tabular}
    }}
\end{table*}
\begin{table}
  \setlength{\belowcaptionskip}{.1cm}
  \setlength{\abovecaptionskip}{0.1cm}
  \caption{Human evaluation on generated texts.}
  \centering
  \renewcommand\arraystretch{.9}
  \setlength{\tabcolsep}{8mm}{
  \scalebox{0.85}{
        \begin{tabular}{c|cc}
           \hline    
           \toprule
              Methods&Usefulness&Fluency\\
           \midrule
           FedJudge-Base       & 3.63 &  4.38 \\
           Base-Client2    & 3.88 &  4.45 \\
              \midrule
              FedJudge-CL   & 3.92 &  4.56 \\
              CL-Client2   & \textbf{4.11} &  \textbf{4.62} \\
           \bottomrule
           \hline
        \end{tabular}
     }
  }   
  \label{table_human}

\end{table}
However, as the number of training epochs increases, we notice that the effectiveness of both CL-Client$\mathcal{E}$ and Base-Client$\mathcal{E}$ improves. Importantly, CL-Client$\mathcal{E}$ almost consistently outperforms Base-Client$\mathcal{E}$, indicating that the continual learning method effectively mitigates data shift problems during the training process.
These findings suggest that the integration of continual learning techniques can enhance the performance of CL-Client$\mathcal{E}$ and alleviate the effects of data shifts, leading to improved results compared to the base approach.

\subsubsection{Human Evaluation (RQ3).}
Table \ref{table1} indicates that both FedJudge-Base and FedJudge-CL exhibit low metrics on the Client2 dataset, prompting a need to explore the underlying reasons. By examining the responses generated by FedJudge and the corresponding ground-truth, it becomes apparent that the ground-truth tends to be shorter, while FedJudge tends to generate longer texts. This discrepancy contributes to FedJudge's poor performance across all metrics. However, upon observation, the quality of text generated by FedJudge appears to be good.
To gain further insights into the impact of FedJudge, a human evaluation is conducted on the text generated by FedJudge in the Client2 dataset. Specifically, we sample 100 examples and ask three annotators who are both good at computer science and law to evaluate the generated text from Usefulness and Fluency. Each metric is scored on a scale from 1 (lowest) to 5 (e.g., 1 and 3). Detailed scoring standards for the human annotators are provided in Table \ref{human}. The experimental results are presented in Table \ref{table_human}.
The results indicate that all models achieve promising scores in terms of Fluency, benefiting from the use of LLMs. Additionally, it is observed that Cl-client2 performs optimally in terms of Usefulness, which aligns with the findings in Table \ref{table1}.

\begin{figure}[!htp]
  \centering 

  \includegraphics[width = 12.cm]{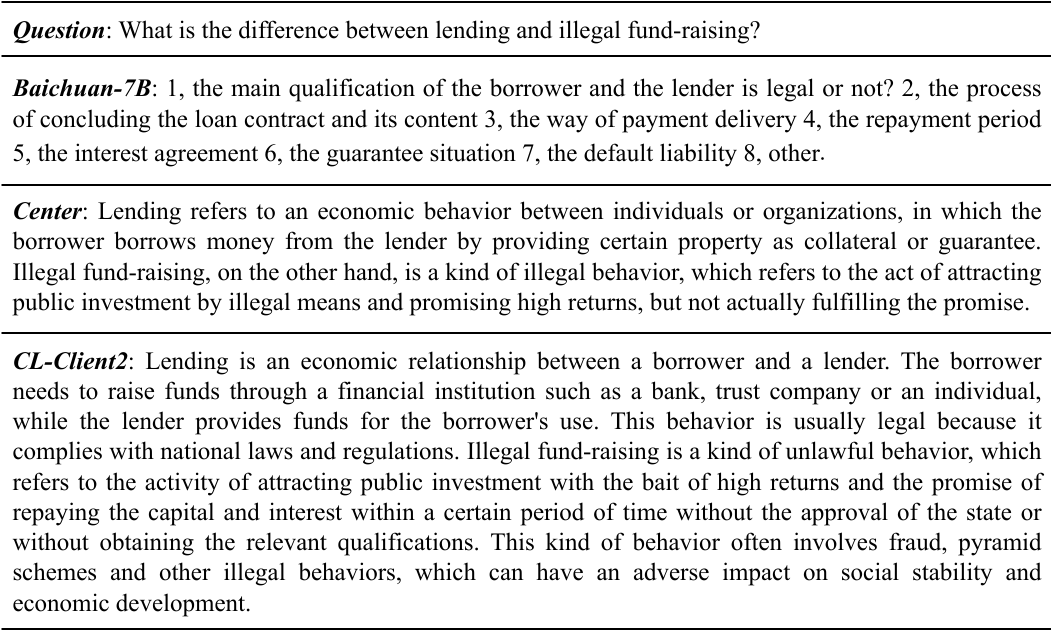}
  
 \caption{
  Visualized  texts with different methods in Legal Consultation  (Client2).
 }
 
 \label{case}
\end{figure}

\subsubsection{Case Study (RQ4).}

\begin{figure}[!htp]
  \centering 

  \includegraphics[width = 12.2cm]{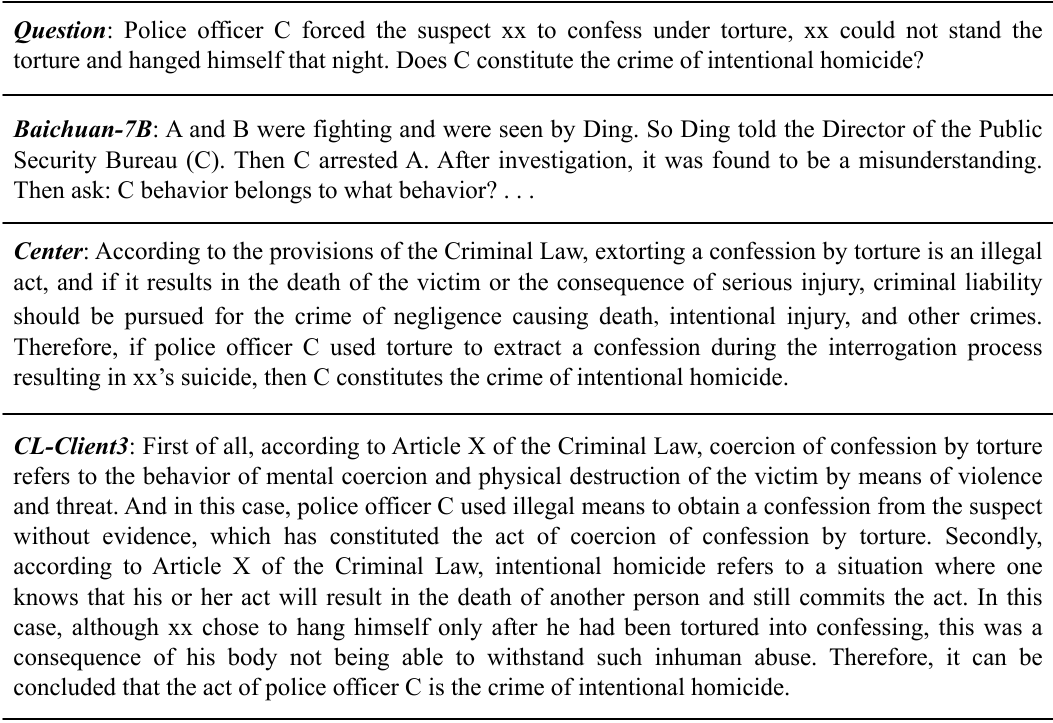}
  
 \caption{
  Visualized texts with different methods in Legal Reasoning (Client3).
 }
 
 \label{case2}
\end{figure}

 In this section, we provide several qualitative analyses of the text generated by our model (CL-Client2 and CL-Client3) and baselines, including Baichuan-7B and Center.
 Specifically, we first visualize an example of Legal Consultation dataset. As shown in Figure \ref{case}, 
 the answer of Baichuan-7B is irrelevant to the question. Both CL-Client2 and Center answer the question correctly, but Cl-Client2 answers more specifically and comprehensively than Center.
 Besides, considering the legal reasoning is a comparatively difficult task,
 we also visualize an example of Legal Reasoning dataset  in Figure~\ref{case2}.
 Similarly, Baichuan-7B response is still irrelevant to the question.
 Although Center correctly reasons the final answer (i.e., \textbf{Police Officer C's action is intentional homicide}), its reasoning process (``\textit{criminal liability should be pursued for the crime of negligence causing death}'')  is wrong.
  In contrast, CL-Client3 correctly answers the question and gives the corresponding reasoning process, which demonstrates the effectiveness of our approach. 
  
\section{Discussion}

\textbf{Ethical Discussion.}
Although FedJudge produces promising outputs and predictions, it is essential to note that these should not be considered as substitutes for professional legal advice. The outputs generated by FedJudge may contain errors or inaccuracies, as it is a machine learning model and not a substitute for the expertise and knowledge provided by legal professionals.
Therefore, if an individual requires legal assistance or professional legal advice, it is highly recommended to seek help from a qualified legal professional, where the results of FedJudge is for reference only.
They possess the necessary expertise, experience, and understanding to provide accurate and reliable legal guidance tailored to specific situations.

\textbf{Limitation.}
While FedJudge has shown promising results on various legal tasks, it currently lacks the ability for multi-round conversations since its training data consists solely of single-round dialogue data. This limitation restricts its application in real-world tasks that involve multi-round conversations. To address this, future work aims to incorporate multi-round conversation data to enhance FedJudge's conversation capabilities.


\section{Conclusions}

In this paper, we focused on how to fine-tune  Large Language Models (LLMs) to the legal  domain under the Federated Learning (FL) setting and further proposed the \textit{first} federated Legal LLMs framework (FedJudge). To be specific,  we developed a parameter-efficient fine-tuning  method to fine-tune FedJudge to achieve efficient training. Additionally, we addressed the challenge of data distribution shifts in FL by incorporating a continual learning method into FedJudge. This approach ensures that important knowledge from the global model is retained and not forgotten during local training.
To validate the effectiveness of FedJudge, we conducted experiments on three real-world datasets. The results of these experiments demonstrated the efficacy of FedJudge in achieving promising performance on various legal tasks.

\textbf{Acknowledgements.}
This research was supported by grants from the National Natural Science Foundation of China (Grants No. 62337001, 623B1020) and the Fundamental Research Funds for the Central Universities.

\bibliographystyle{splncs04}

\bibliography{ref}

\end{document}